\documentclass[preprint,12pt]{elsarticle}

\usepackage{graphicx}
\usepackage{subfigure}

\usepackage[justification=centering]{caption}
\usepackage{algpseudocode}
\usepackage{algorithmicx,algorithm}
\usepackage{hyperref}
\usepackage{marvosym}
\usepackage{colortbl}
\pdfoutput=1

\usepackage{amssymb}

\usepackage{lineno}
\usepackage{appendix}
\usepackage{url}

\biboptions{sort&compress}

\usepackage{geometry}
\usepackage{color}
\usepackage{indentfirst}

\usepackage{amsmath}

\usepackage{multirow}

\usepackage{bm}

\newtheorem{thm}{Theorem}[section]
\newtheorem{lem}{Lemma}[section]
\newdefinition{rmk}{Remark}[section]
\newdefinition{defn}{Definition}[section]
\newdefinition{cor}{Corollary}[section]
\newdefinition{asm}{Assumption}[section]
\newproof{pf}{Proof}

\journal{Neurocomputing}

\everymath{\displaystyle}

\begin{document}

\begin{frontmatter}



\title{A decreasing scaling transition scheme from Adam to SGD \tnoteref{t1}}

\tnotetext[t1]{This work was funded in part by the National Natural Science Foundation of
China (Nos. 62176051, 61671099), in part by National Key R\&D Program of China (No.2020YFA0714102), and in part by the Fundamental Research Funds for the
Central Universities of China (No. 2412020FZ024).}


\author[a]{Kun Zeng}
\ead{zki@163.com}
\author[b]{Jinlan Liu}
\author[a]{Zhixia Jiang\corref{cor1}}
\ead{zhixia\_jiang@126.com}
\author[b]{Dongpo Xu\corref{cor1}}
\ead{xudp100@nenu.edu.cn}
\cortext[cor1]{Corresponding authors: Zhixia Jiang and Dongpo Xu.}



\address[a]{School of Mathematics and Statistics, Changchun University of Science and Technology, Changchun 130022, China}
\address[b]{School of Mathematics and Statistics, Northeast Normal University, Changchun 130024, China}


\begin{abstract}
Adaptive gradient algorithm (AdaGrad) and its variants, such as RMSProp, Adam, AMSGrad, etc, have been widely used in deep learning.  Although these algorithms are faster in the early phase of training, their generalization performance is often not as good as stochastic gradient descent (SGD). Hence, a trade-off method of transforming Adam to SGD after a certain iteration to gain the merits of both algorithms is theoretically and practically significant.
To that end,  we propose a decreasing scaling transition scheme to achieve a smooth and stable transition from Adam to SGD, which is called  DSTAdam. The convergence of the proposed DSTAdam is also proved in an online convex setting.  Finally, the effectiveness of the DSTAdam is verified on the CIFAR-10/100 datasets. Our implementation is available at: \href{https://github.com/kunzeng/DSTAdam}{https://github.com/kunzeng/DSTAdam}.
\end{abstract}


\end{frontmatter}


\section{Introduction}
\label{S:1}

Deep neural networks (DNN) have achieved great success in many fields, such as computer vision\cite{krizhevsky2012imagenet}, natural language translation\cite{wu2016google}, medical treatment\cite{rahimiazghadi2020hardware} and transportation\cite{kato2016deep}. Many popular neural network models have been proposed, such as fully connected neural network\cite{rumelhart1986learning}, LeNet\cite{lecun1989handwritten}, LSTM\cite{hochreiter1997long}, ResNet\cite{he2016deep}, DensNet\cite{huang2017densely}, etc.
During the training of these models, the optimization algorithms always play a significant role.
One of the most dominant algorithms is stochastic gradient descent (SGD) \cite{RobbinsMonro1951} with a simple mathematical form and low computational complexity. However, SGD only depends on the current stochastic gradient and is not stable enough in the training process.
Polyak et al.\cite{polyak1964some} added a momentum term into SGD and employed a constant learning rate to scale the gradient in all directions.
Duchi et al.\cite{duchi2011adaptive} proposed an adaptive gradient (AdaGrad) algorithm with the second moment as the adaptive learning rate.
Furthermore, there are many variants of AdaGrad, such as RMSProp\cite{HintonRMSProp}, AdaDelta\cite{zeiler2012adadelta} and Adam \cite{kingma2014adam}.
Adam is a combination of AdaGrad and RMSProp, which has been widely used into various deep neural networks. Reddi et al.\cite{reddi2019convergence} found that Adam could not converge for a convex optimization problem, and proposed AMSGrad to make up for this deficiency.

Another methodology is to transform Adam to SGD in the later stages of training \cite{keskar2017improving,luo2019adaptive,liang2020new,yang2021adadb,lu2020distributed}, which can have the advantages of rapid convergence of Adam and good generalization of SGD.
In particular, we would like to point out that Keskar et al.\cite{keskar2017improving} manually switched Adam to SGD after a certain moment in the training process, which satisfies the conditions $ k > 1$ and $\left| {{\lambda _k}/\left( {1 - \beta _2^k} \right) - {\gamma _k}} \right| < \varepsilon $. However, it is difficult to obtain the optimal  transition moment for the ``hard'' transition in practice.
Luo et al.\cite{luo2019adaptive} investigated the extreme learning rates of Adam and
proposed an adaptive gradient algorithm with two dynamic bound functions (named AdaBound) to achieve a smooth transition from Adam to SGD.
However, AdaBound is data-independent on the chosen bound functions and  may hurt the generalization  performanc\cite{yang2021adadb}.
Therefore, how to achieve a better transition from Adam to SGD still needs more and further investigation.

In this paper, we firstly introduce a general transition framework and propose a decreasing scaling scheme to achieve a smooth and stable transition from Adam to SGD.
Through the scaling, the adaptability of Adam is retained to some extent during the transition process.
In the training stage of SGD, we employ linear decreasing learning rate instead of a constant learning rate to obtain better experimental results.
Furthermore, we prove that DSTAdam has a regret bound $\mathcal{O}( {\sqrt T } )$ in the online convex setting.
Experimental results show that the proposed DSTAdam has faster convergence speed and better generalization performance.

\section{Preliminaries}
\label{S:2}
\noindent\textbf {\fontsize{12pt}{0} \selectfont Notation.} For $x \in {\mathbb{R}^d}$, we use ${x_i}$ to denote the $i$-th component of vector $x$. ${\left\| x \right\|_2}$ and ${\left\| x \right\|_\infty }$ respectively denote the
${l_2}$-norm and ${l_\infty }$-norm of $x$. For any two vectors $x,y \in {\mathbb{R}^d}$, $\left\langle {x,y} \right\rangle $ signifies the inner product. $xy$ and $x/y$ denote element-wise product and element-wise division respectively. $\mathcal{S}_{+}^{d}$ means the set of all positive definite $d \times d$ matrices. For a vector $a \in \mathbb{R}^{d}$ and a positive definite matrix $M \in \mathbb{R}^{d \times d}$, we use $a / M$ to denote $M^{-1} a$ and $\sqrt{M}$ to denote $M^{1 / 2}$. The projection operation $\Pi_{\mathcal{F}, M}(y)$ for $M \in \mathcal{S}_{+}^{d}$
 is defined as $\arg \min _{x \in \mathcal{F}}\left\|M^{1 / 2}(x-y)\right\|$ for $y \in \mathbb{R}^{d}$. The set $\{1,2,\cdots,T\}$ is denoted as $[T]$ .

\begin{algorithm}[htbp]
\caption{Generic framework of transition methods}\label{alg:sgd}\label{alg:1}
{\bf Input:}  $\theta_1 \in \mathcal{F}$, step size $\{\eta_t\}_{t=1}^T$,  sequence of functions $\{\phi_{t},\psi_{t}\}_{t=1}^T$, bound functions $\eta_l$ and $\eta_u$.
\begin{algorithmic}[1]
\For{$t=1$ to $T$}
    \State Generate a stochastic gradient ${g_t} = \nabla {f_t}({\theta _t})$
    \State $m_{t}=\phi_{t}\left(g_{1}, \ldots, g_{t}\right) \text \ and \ v_{t}=\psi_{t}\left(g_{1}, \ldots, g_{t}\right)$
    \State ${\eta _t} = \text{Clip}\left( {\alpha /\sqrt {{v_t}}, {\eta _l}\left( t \right), {\eta _u}\left( t \right)} \right)$
    \State  $\theta_{t+1}=\Pi_{\mathcal{F}, \operatorname{diag}\left(\eta_{t}^{-1}\right)}\left(\theta_{t}-\eta_{t} m_{t}\right)$
\EndFor
\end{algorithmic}
\end{algorithm}

\noindent\textbf {\fontsize{12pt}{0} \selectfont Optimization problem.} In the online learning framework \cite{zinkevich2003online}, the sequence of convex loss function is defined as ${f_1}\left( {{\theta _1}} \right),{f_2}\left( {{\theta _2}} \right),...,{f_T}\left( {{\theta _T}} \right)$, where ${\theta _t}$  is the parameter that needs to be optimized. Then, the regret of the problem is
defined as the sum of the difference between each iteration loss and the optimal loss \cite{zhou2018adashift}, that is
\begin{equation}
R(T) = \sum\limits_{t = 1}^T {\left( {{f_t}({\theta_t}) - {f_t}\left( {{\theta^*}} \right)} \right)},
\end{equation}
 where ${\theta^*} = \arg\min_{\theta\in \mathcal F}\sum\nolimits_{t=1}^T {f_t}( \theta )$. Throughout this paper, we assume that the feasible
set $\mathcal F$ has bounded diameter ${D_\infty }$ such that $\left\| {x - y} \right\| \le {D_\infty }$, $\forall x,y \in \mathcal F$.

\noindent\textbf {\fontsize{12pt}{0} \selectfont A generic overview of transition methods.}
Following the clipping operation of \cite{luo2019adaptive}, we now provide a generic framework of transition methods in Algorithm~\ref{alg:1}, which encapsulates many popular transition methods so that we can better understand the differences between different transition methods.
In Algorithm 1, $\phi_t$ and $\psi_t$ denote the "averaging" functions which have not been specified, $\eta _l(t)$ and $\eta _u(t)$ denote the lower
and upper bound functions. Based on such a framework, we have listed most of the existing bound functions in Table~\ref{table:1}.

\begin{table}[!ht]
\centering
\caption{Some bound functions}\label{table:1}
\begin{tabular}{|c|c|c|}
\hline  $\eta_{l}(t)$ & $\eta_{u}(t)$ &References  \\
\hline  $\alpha^{*}$ & $\alpha^{*}$ &\cite{keskar2017improving} \\
\hline  $\alpha^{*}\left(1-\frac{1}{ \left(1-\beta_{2}\right) t+1}\right) $ & $\alpha^{*}\left(1+\frac{1}{\left(1-\beta_{2}\right) t}\right) $ & \cite{luo2019adaptive,liang2020new} \\
\hline  $\alpha^{*}$ & $\alpha^{*}+\frac{\left|m_{t}\right|}{\max \left\{\left|m_{t}\right|\right\} \cdot(\gamma t)}$&\cite{yang2021adadb}  \\
\hline $\alpha^{*}\cdot\frac{t}{T}$ & $\alpha^{*} +\frac{1}{\left(1-\beta_{2}\right) \cdot t}-\frac{1}{\left(1-\beta_{2}\right) \cdot T}$ &\cite{lu2020distributed} \\
\hline
\end{tabular}
\end{table}
From Table~\ref{table:1}, we can see that most of the existing transition methods from adaptive gradient descent to stochastic gradient descent are improvement on the lower and upper bound functions of clipping operation,
but there are few studies on such transition operation.

\section{DSTAdam algorithm}
\label{S:3}
In this section, we propose a new transition method from Adam to SGD.
For the SGD algorithm, we make use of linear decreasing learning rate to accelerate the training speed in the early stage and stabilize the convergence in the later stage.
A scaling function $\rho_t$ is introduced to gradually scale the adaptive learning rate to the linear decreasing learning rate, which retains the adaptability of Adam to a certain extent, and achieves a smooth and stable transition from Adam to SGD. The specific algorithm is described in Algorithm~\ref{alg:DSTAdam}.

\begin{algorithm}[htbp]
\caption{\textbf{D}ecreasing \textbf{S}caling \textbf{T}ransition from \textbf{Adam} to SGD (DSTAdam) Algorithm}\label{alg:DSTAdam}
{\bf Input:}  $\theta_1 \in \mathcal{F}$, initial step size $\alpha$, $r_l$, $r_u$, $\{{\rho_t}\}$, $\{{\beta _{1t}}\}$ and $\{{\beta _{2t}}\}$. \\
\textbf{Initialize:} $m_{0}=0, v_{0}=0.$
\begin{algorithmic}[1]
\For{$t=1$ to $T$}
    \State ${g_t} = \nabla {f_t}({\theta _t})$
    \State ${m_t} = {\beta _{1t}}{m_{t - 1}} + (1 - {\beta _{1t}}){g_t}$
    \State ${v_t} = {\beta _{2t}}{v_{t - 1}} + (1 - {\beta _{2t}})g_t^2$
    \State ${r_t} = \left( {r_u - r_l} \right)(1 - t/T) + r_l$
    \State ${\hat \eta_t} = \rho_t\left( {\alpha/\sqrt {{v_t}}  - {r_t}} \right) + {r_t}$ {and} $\eta_{t}=\hat{\eta}_{t} / \sqrt{t}$ \label{step:6}
    \State $\theta_{t+1}=\Pi_{\mathcal{F}, \operatorname{diag}\left(\eta_{t}^{-1}\right)}\left(\theta_{t}-\eta_{t} m_{t}\right)$
\EndFor
\end{algorithmic}
\end{algorithm}

For the sake of clarity, we omit the bias correction operations of the first and second moments in DSTAdam. Following the hyperparameter setting of Adam \cite{kingma2014adam}, the hyperparameters of DSTAdam are taken as $\alpha = 0.001$, ${\beta _{1t}} = 0.9$ and ${\beta _{2t}} = 0.999$.
In a similar manner as \cite{luo2019adaptive}, we set $\eta_{t}=\hat{\eta}_{t} / \sqrt{t}$ for theoretical analysis, but $\eta_{t}=\hat{\eta}_{t}$ is more effective in practice.

\begin{figure}[htbp]
\centering
\subfigure[Learning rates of Adam.]{
\begin{minipage}[t]{0.3\linewidth}
\centering
\includegraphics[width=2.0in]{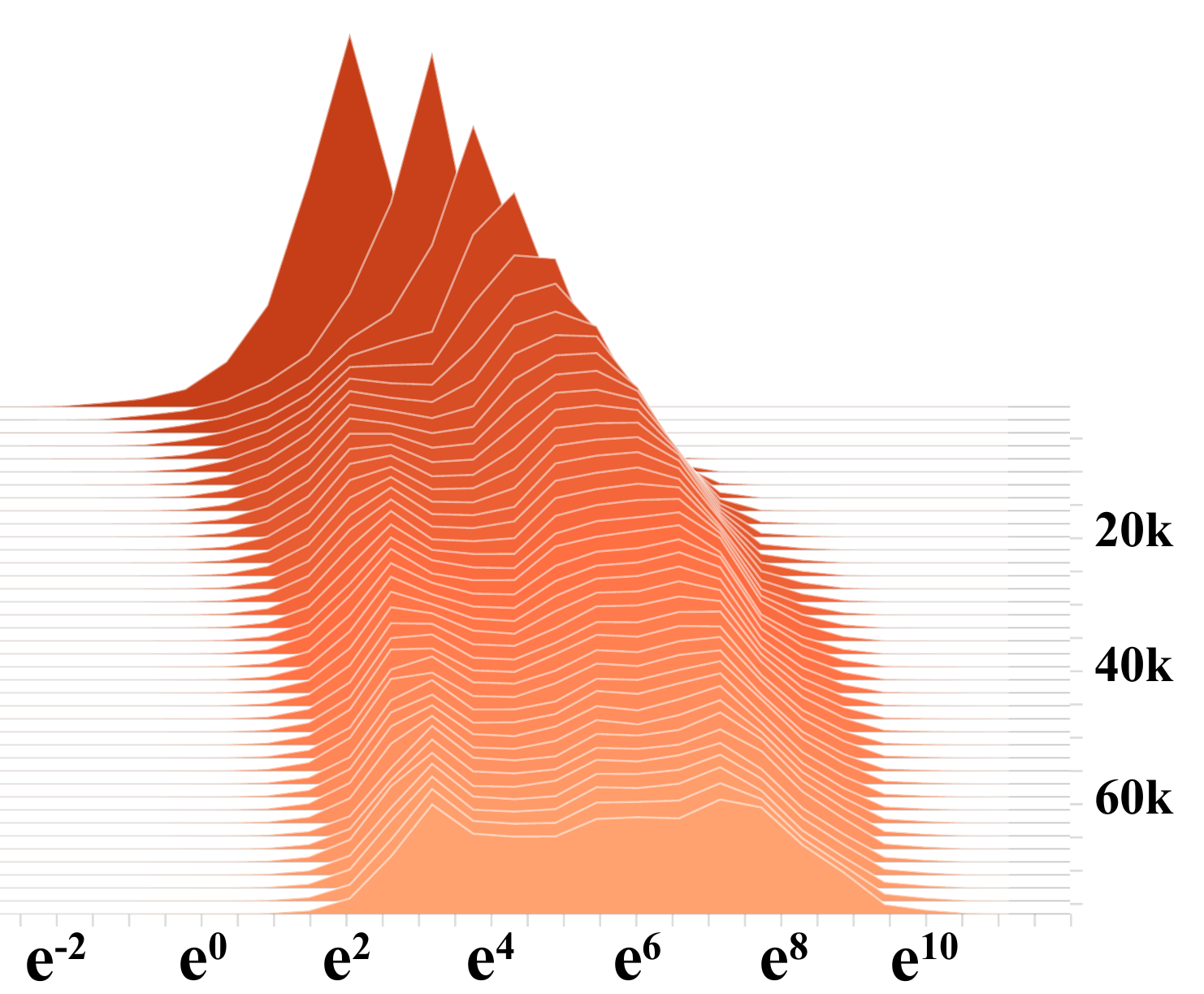}
\hspace{0.1in}
\end{minipage}%
}%
\subfigure[Learning rates of AdaBound.]{
\begin{minipage}[t]{0.3\linewidth}
\centering
\includegraphics[width=2.0in]{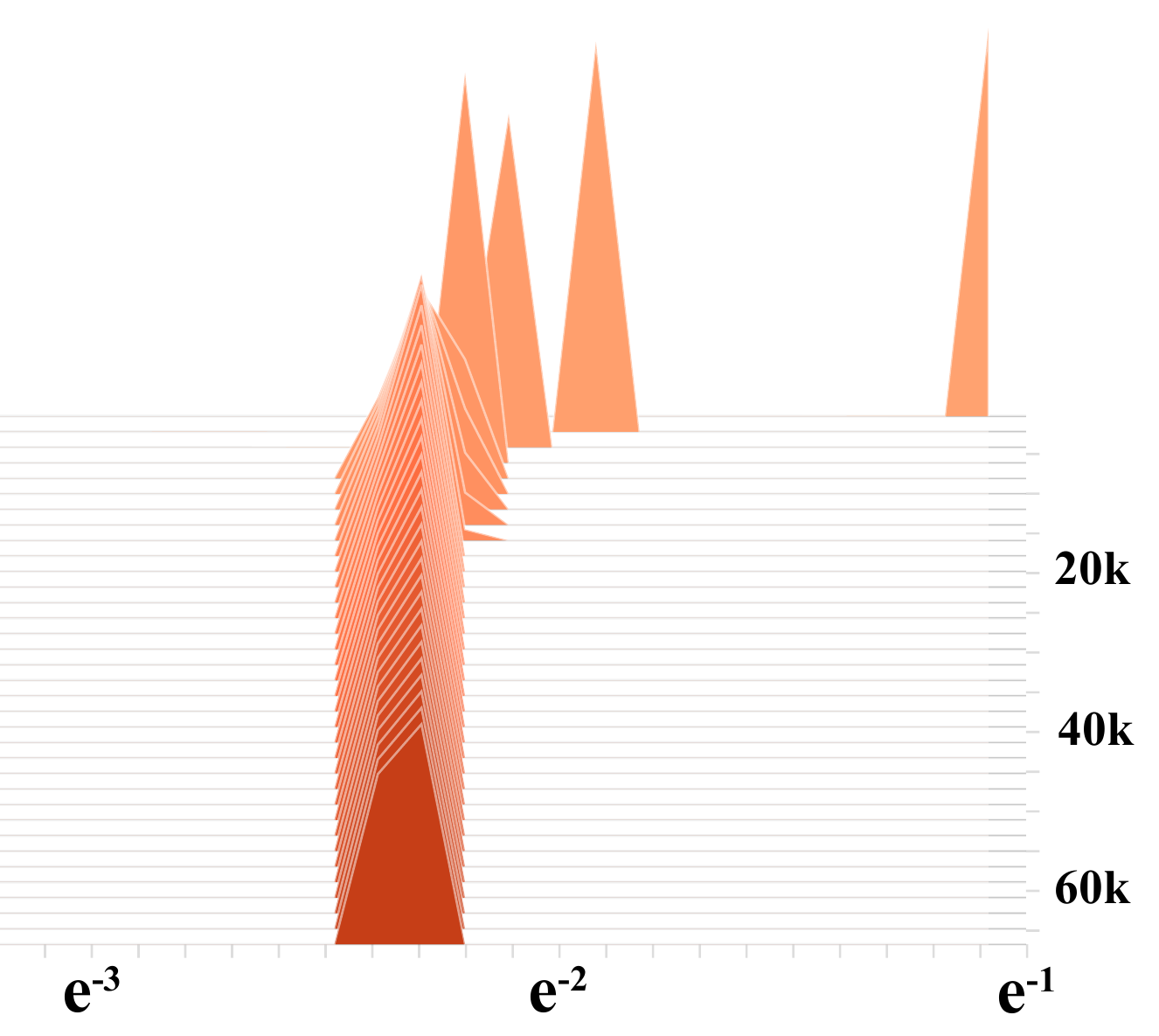}
\hspace{0.1in}
\end{minipage}%
}%
\subfigure[Learning rates of DSTAdam.]{
\begin{minipage}[t]{0.3\linewidth}
\centering
\includegraphics[width=2.0in]{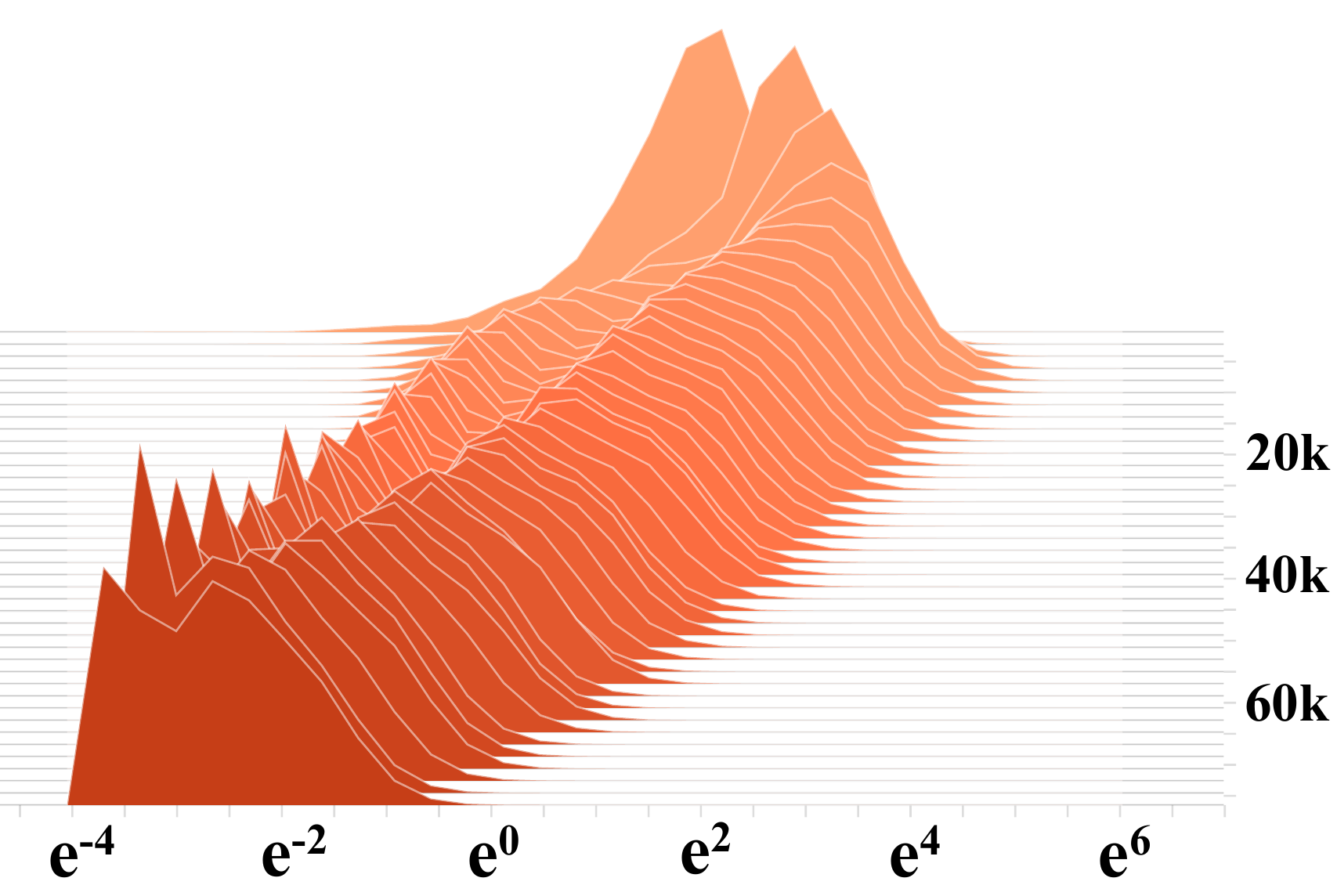}
\hspace{0.1in}
\end{minipage}
}%
\caption{The schematic diagrams of the learning rates distribution of the Adam, AdaBound and DSTAdam}\label{fig:1}
\end{figure}

\begin{rmk}
It should be noted that the transition steps 5-6 of Algorithm~\ref{alg:DSTAdam} is more smooth and stable through the scaling operation instead of the clipping operation. This technology effectively reduces the dependence of AdaBound on the bound functions and restores the element-to-element operation of the effective learning rates. The schematic diagrams\footnote{This diagram is generated by TensorBoardX: https://github.com/lanpa/tensorboardX.} of the learning rates distribution of the training process of Adam, AdaBound and DSTAdam for ResNet-18 on CIFAR-10 are shown in Fig.~\ref{fig:1}, where the $x$-axis is scaled by log function and the $y$-axis denotes the number of iterations. We can see that $i$) the learning rates of Adam are widely distributed, and there are a lot of extreme learning rates in the later stage of training, $ii$) the learning rates of the AdaBound quickly shrink to a small interval after a few iterations due to the clipping operation of two bound functions of AdaBound,
and $iii$) the learning rates of the DSTAdam are similar to that of Adam in the early stage of training,
and its learning rates gradually transform to a small range with the number of iterations.
In contrast to Adam and AdaBound, the transition process of DSTAdam eliminates extreme learning rates while retaining its adaptability to a certain extent.

\end{rmk}

\section{Convergence Analysis of DSTAdam}
\label{S:4}
Under the similar conditions as \cite{kingma2014adam,luo2019adaptive,reddi2019convergence,bock2018improvement}, we obtain the following convergence theorem for the DSTAdam, and its proof is given in the Appendix~\ref{app:A}.

\begin{thm}\label{thm:main}
Let $\left\{ {{\theta _t}} \right\}$ and $\left\{ {{\eta_t}} \right\}$ be the sequence generated by DSTAdam, ${\rho _t} \le \rho $, $\rho \in \left( {0,1} \right)$, $\eta_{t}=\hat{\eta}_t / \sqrt{t}$, $r_l \le r_u$, ${\beta _{1t}} \le {\beta _1}$ for all $t \in \left[ T \right]$, ${\beta _1} \in \left[ {0,1} \right)$. Assume that $\left\|x-y\right\|_{\infty} \leq D_{\infty}$, $\forall x,y\in \mathcal F$ and $\left\|\nabla f_{t}(\theta)\right\|_{\infty} \leq G_{\infty}$, $\forall t \in \left[ T \right]$ and $\forall \theta \in \mathcal F$. Furthermore, let $\left\{\beta_{2 t}\right\}$ be such that the following conditions are satisfied:
\[\begin{aligned}
&\textbf {(C1). } \sqrt{t \sum_{j=1}^{t} \prod_{k=1}^{t-j} \beta_{2(t-k+1)}\left(1-\beta_{2 j}\right) g_{j, i}^{2}} \geq \frac{1}{\zeta} \sqrt{\sum_{j=1}^{t} g_{j, i}^{2}} \text { for some } \frac{1}{\zeta}>0 \text { and all } t \in[T], j \in[d].\\
&\textbf {(C2). } \sqrt{t} \hat{\eta}_{t, i}^{-1} \geq \sqrt{t-1} \hat{\eta}_{t-1, i}^{-1} \text { for all } t \in\{2, \ldots, T\} \text { and } i \in[d].
\end{aligned}\]

Then, we have the following bound on the regret
\[\begin{aligned}
R(T) \leq & \frac{\sqrt{T} D_{\infty}^{2}}{2\left(1-\beta_{1}\right)} \sum_{i=1}^{d} \hat{\eta}_{T, i}^{-1}+\frac{D_{\infty}^{2}}{2\left(1-\beta_{1}\right)} \sum_{t=1}^{T} \sum_{i=1}^{d} \frac{\beta_{1 t} \sqrt{t}}{\hat{\eta}_{t, i}} \\
&+\frac{2 \alpha \rho \zeta}{\left(1-\beta_{1}\right)^{3}} \sum_{i=1}^{d}\left\|g_{1: T, i}\right\|_{2}+\frac{r_{u} \sqrt{1+\log T}}{\left(1-\beta_{1}\right)^{3}} \sum_{i=1}^{d}\left\|g_{1: T, i}^{2}\right\|_{2} .
\end{aligned}\]
\end{thm}

The following two results fall immediate consequences of the theorem \ref{thm:main}
\begin{cor}\label{cor:4.1}
Suppose that ${\beta _{1t}} = {\beta _1}{\lambda ^{t - 1}}$ with $\lambda  \in \left( {0,1} \right)$ in Theorem \ref{thm:main}, then we have
\[\begin{aligned}
R(T) \leq & \frac{\sqrt{T} D_{\infty}^{2}}{2\left(1-\beta_{1}\right)} \sum_{i=1}^{d} \hat{\eta}_{T, i}^{-1}+\frac{d D_{\infty}^{2}}{2 r_{l}(1-\rho)(1-\lambda)^{2}\left(1-\beta_{1}\right)} \\
&+\frac{2 \alpha \rho \zeta}{\left(1-\beta_{1}\right)^{3}} \sum_{i=1}^{d}\left\|g_{1: T, i}\right\|_{2}+\frac{r_{u} \sqrt{1+\log T}}{\left(1-\beta_{1}\right)^{3}} \sum_{i=1}^{d}\left\|g_{1: T, i}^{2}\right\|_{2}.
\end{aligned}\]
\end{cor}

\begin{cor}\label{cor:4.2}
Suppose that ${\beta _{1t}} = {\beta _1}/t$ in Theorem \ref{thm:main}, then we have
\[\begin{aligned}
R(T) \leq & \frac{\sqrt{T} D_{\infty}^{2}}{2\left(1-\beta_{1}\right)} \sum_{i=1}^{d} \hat{\eta}_{T, i}^{-1}+\frac{d D_{\infty}^{2} \sqrt{T}}{r_{l}(1-\rho)\left(1-\beta_{1}\right)} \\
&+\frac{2 \alpha \rho \zeta}{\left(1-\beta_{1}\right)^{3}} \sum_{i=1}^{d}\left\|g_{1: T, i}\right\|_{2}+\frac{r_{u} \sqrt{1+\log T}}{\left(1-\beta_{1}\right)^{3}} \sum_{i=1}^{d}\left\|g_{1: T, i}^{2}\right\|_{2}.
\end{aligned}\]
\end{cor}

\begin{rmk}
Note that $\hat{\eta}_{T, i}^{-1}\leq\frac{1}{r_{l}(1-\rho)}$ in \eqref{eq:etbound}, ${{{\left\| {{g_{1:T,i}}} \right\|}_2}}  \le \sqrt T {G_\infty }$ and $ {{{\left\| {g_{1:T,i}^2} \right\|}_2}}  \le \sqrt T G_\infty ^2$, we can show DSTAdam has ${\mathcal O}(\sqrt T)$ regret bound for two  momentum decay cases of ${\beta _{1t}} = {\beta _1}{\lambda ^{t - 1}}$ with $\lambda  \in \left( {0,1} \right)$ and ${\beta _{1t}} = {\beta _1}/t$.
\end{rmk}

\section{Experiments}
\label{S:5}

In this section, we use ResNet-18 \cite{he2016deep} to empirically study the performance of DSTAdam on the CIFAR datasets \cite{krizhevsky2009learning} compared with other optimizers, including SGD with momentum(SGDM), Adam and  AdaBound.
In particular, the scaling function in DSTAdam is recommended to be an exponential decay function $\rho_t = \rho^t$, $\rho\in(0,1)$ in this experiment.
The same random seed is set in each experiment for a fair comparison. The computational setup and hyperparameter setting are given in in the Table~\ref{table:compu} and Table~\ref{table:hyper}.
\begin{table}[!htbp]
\caption{Computational setup}\label{table:compu}
\centering
	\begin{tabular}{|ll|ll|}
	\hline OS & CentOS 8.3 & Dataset storage location & OS File system \\
	\hline Processor-CPU & Intel Core i7-6500U & CPU-Memory & $32.0 \mathrm{~GB}$ \\
	\hline Processor-GPU & Quadro P600 & GPU-Memory & $2.0 \mathrm{~GB}$ \\
	\hline Programming language & Python $3.7 .10$ & Development framework & Pytorch $1.7 .0$ \\
	\hline
	\end{tabular}
\end{table}

\begin{table}[H]
\caption{The hyperparameter setting for ResNet-18 on CIFAR-10/100}\label{table:hyper}
\centering
\begin{tabular}{|c|c|c|c|c|c|c|c|c|c|}
\hline & Epoch & $\operatorname{LR}$ & $\beta_{1t}$ & $\beta_{2t}$  & Batch Size & $\alpha^{*}$ & $\rho$ & $r_{u}$ & $r_{l}$ \\
\hline SGDM & 200 & $0.1$ & $0.9$ & $-$ & 128 & $-$ & $-$ & $-$ & $-$ \\
\hline Adam & 200 & $0.001$ & $0.9$ & $0.999$  & 128 & $-$ & $-$ & $-$ & $-$ \\
\hline AdaBound & 200 & $0.001$ & $0.9$ & $0.999$ & 128 & $0.1$ & $-$ & $-$ & $-$ \\
\hline DSTAdam & 200 & $0.001$ & $0.9$ & $0.999$  & 128 & $-$ & $0.999764$ & $5$ & $0.005$ \\
\hline
\end{tabular}
\end{table}

\noindent\textbf {\fontsize{12pt}{0} \selectfont Hyperparameter tuning.}
The common hyperparameter values of Adam, AdaBound and DSTAdam follow from  the default hyperparameters of Adam.
We perform a grid search to select the hyperparameters $r_u\in\left\{0.1, 0.5, 1, 5, 10\right\}$, $r_l\in\left\{0.001 , 0.005, 0.01,0.1 \right\}$.
In this experiment, we recommend taking $r_u = 5$ and $r_l = 0.005$ for CIFAR-10 and CIFAR-100 datasets.
Next, we provide an empirical formula for choosing the transition factor $\rho$, that is ${\rho ^T}=10^{-8}$, where $T$ is number of iterations\footnote{The number of iterations is equal to the training sample size divided by the batch size, rounded up and then multiplied by the number of epochs.}. In our experiments, the training sample size is 50000, the batch size is 128, and the number of epochs is 200, then $T=78200$ and $\rho\approx0.999764$.\\

\begin{figure}[htbp]
  \centering
  \subfigure[Training loss]{\includegraphics[width=3in]{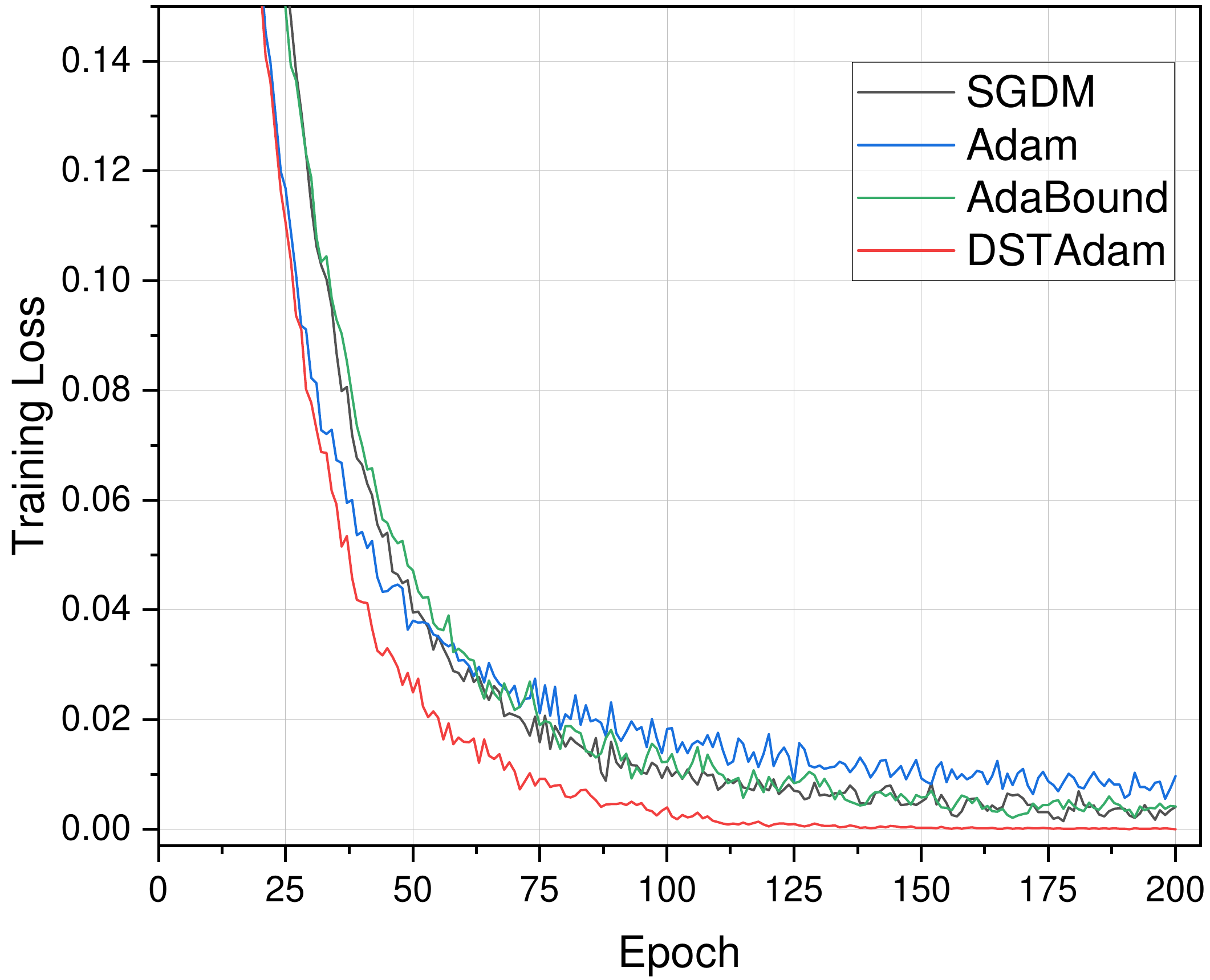}}
  \hspace{0.1in}
  \subfigure[Test accuracy]{\includegraphics[width=3in]{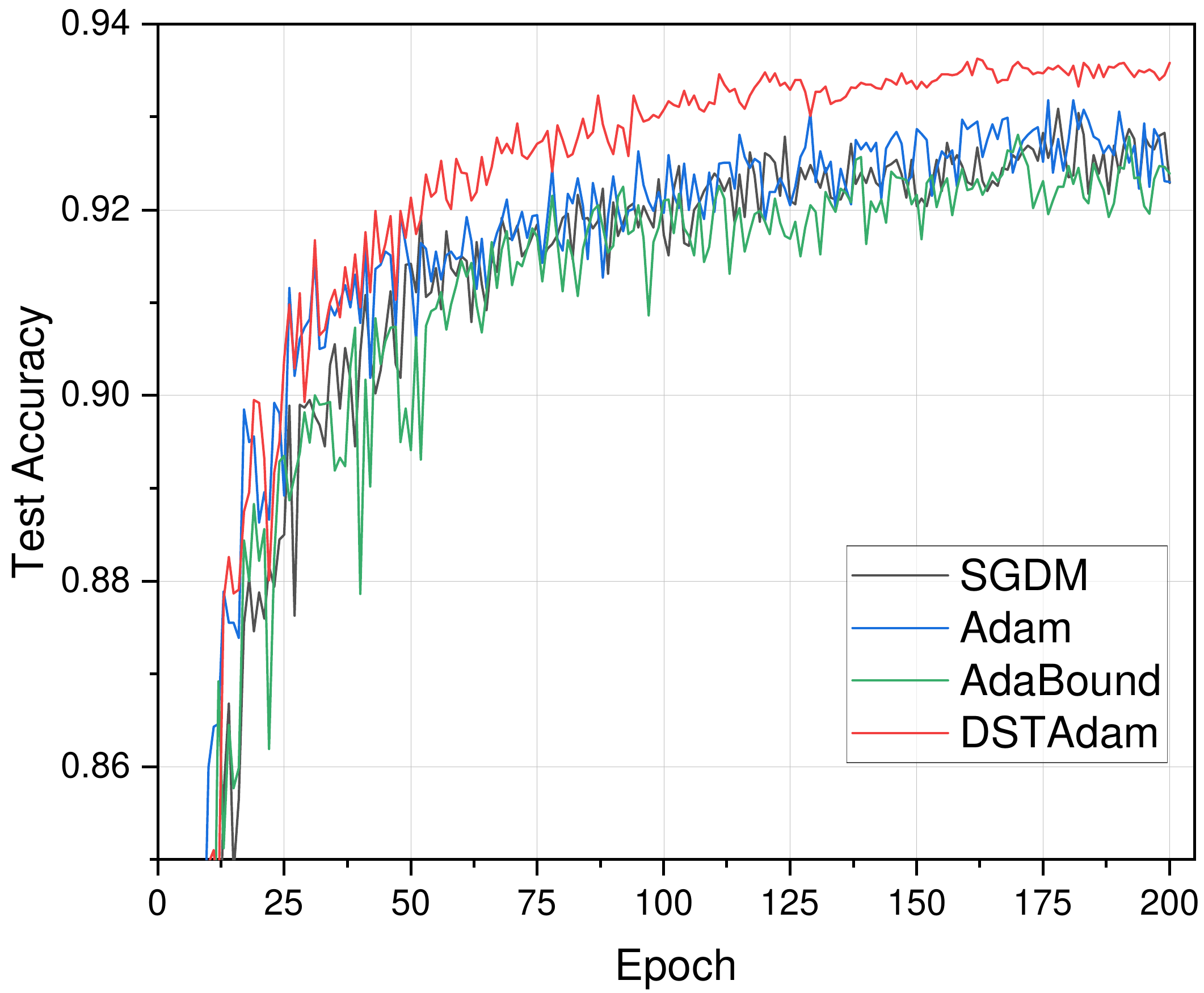}}
  \caption{Training loss and test accuracy for ResNet-18 on CIFAR-10.}
  \label{fig:cifar10}
\end{figure}

\noindent\textbf {\fontsize{12pt}{0} \selectfont Results on CIFAR-10 dataset.} We firstly test the performance of the proposed DSTAdam on the CIFAR-10 dataset, which consists of 60,000 32$\times$32 RGB color images drawn from 10 categories.
Each category contains 5,000 training samples and 1,000 test samples.
The training loss and the test accuracy of the four methods versus the number of epochs are plotted in Figure~\ref{fig:cifar10}.
From subfigure 2(a), we see that DSTAdam has faster convergence speed and lower steady-state loss than other algorithms.
Figure 2(b) shows that the test accuracy of DSTAdam is also better than other algorithms, with an improvement of about 1\% at the end of training.
In addition, the loss and accuracy curves of DSTAdam are more smooth and stable, which reflect the effect of the decreasing scaling transition.\\

\begin{figure}[htbp]
  \centering
  \subfigure[Training loss]{\includegraphics[width=3in]{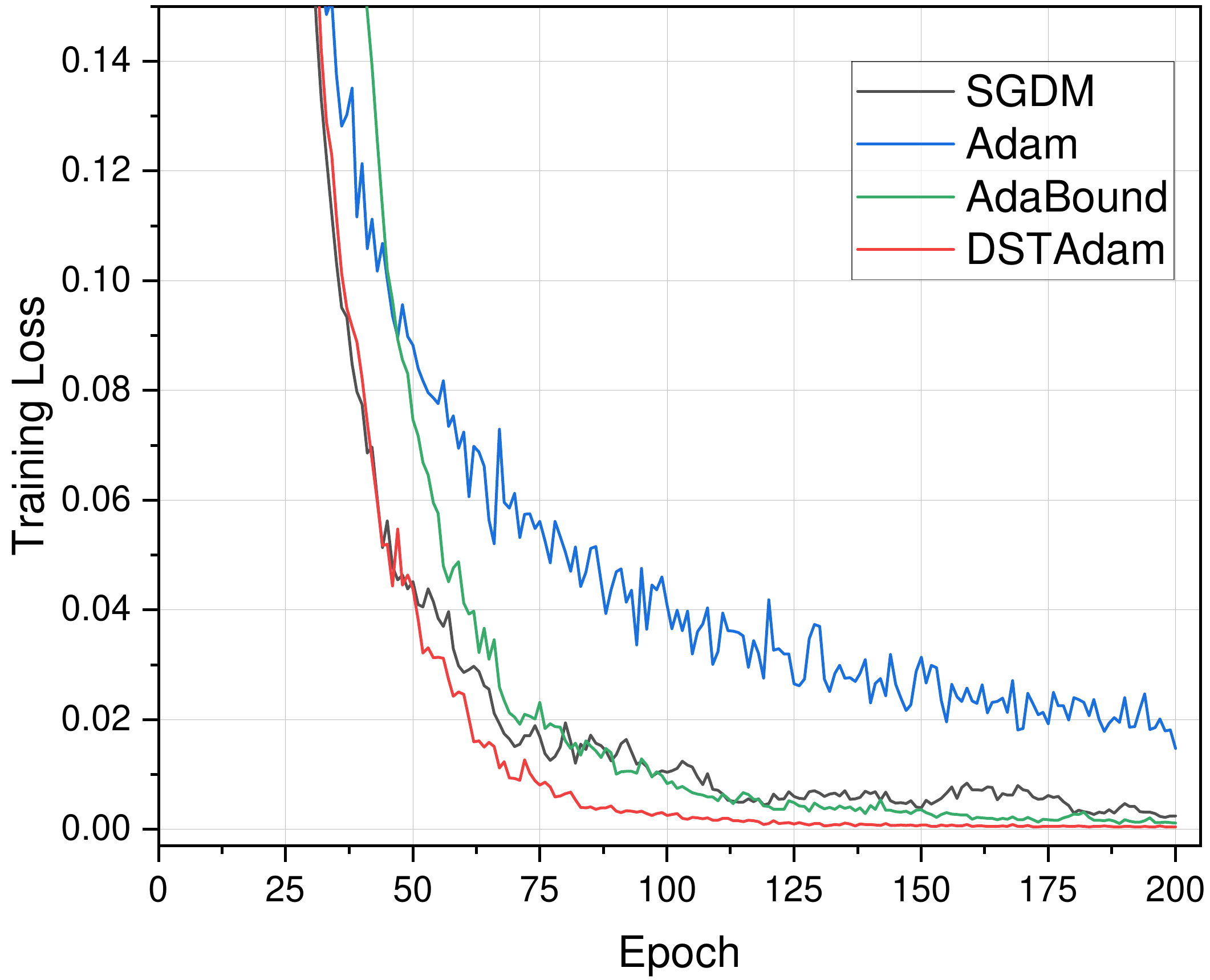}}
  \hspace{0.1in}
  \subfigure[Test accuracy]{\includegraphics[width=3in]{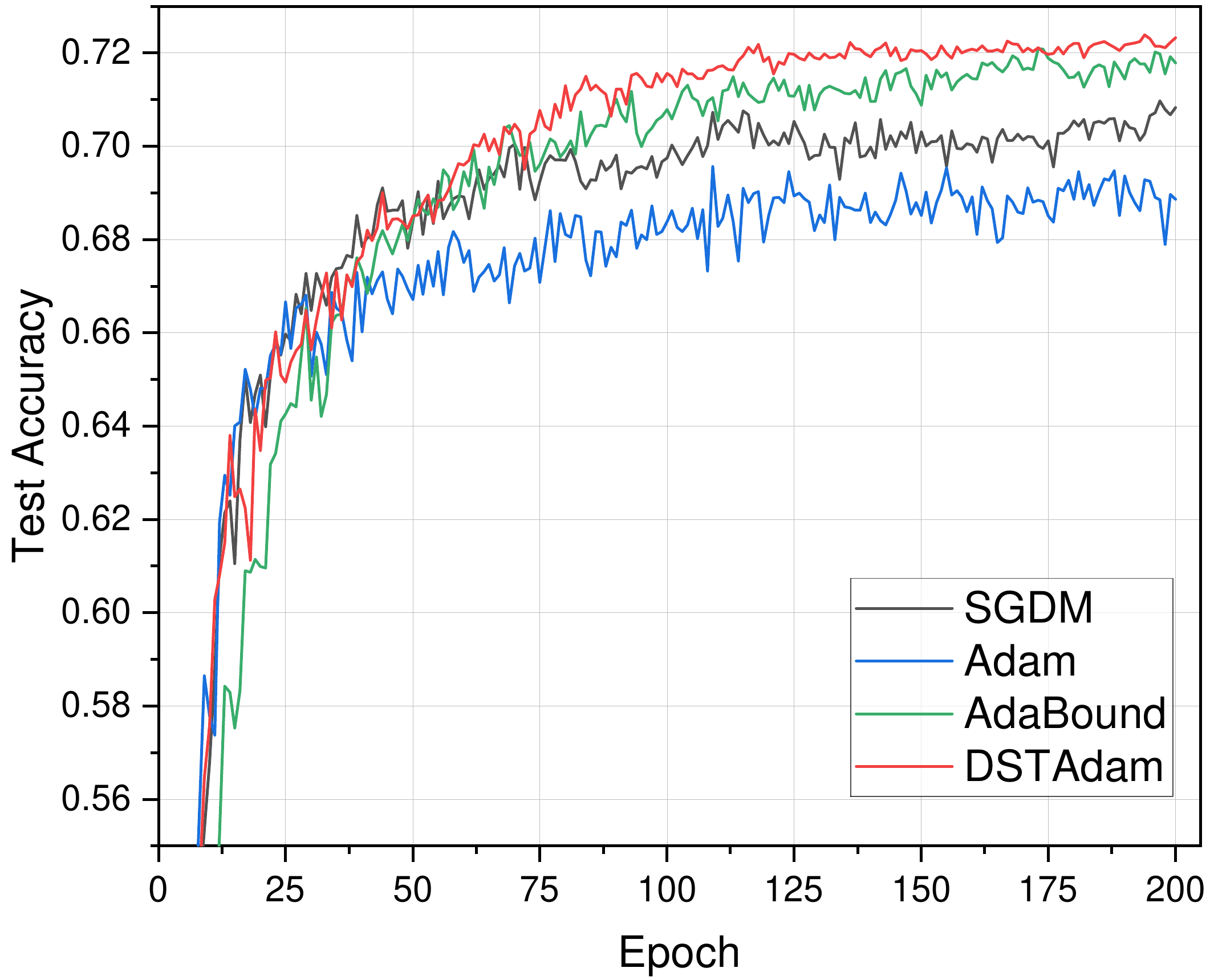}}
  \caption{Training loss and test accuracy for ResNet-18 on CIFAR-100.}
  \label{fig:cifar100}
\end{figure}

\noindent\textbf {\fontsize{12pt}{0} \selectfont Results on CIFAR-100 dataset.} We next perform experiments on the CIFAR-100 dataset,
which contains 60,000 32$\times$32 color images in 100 classes. Each class has 500 training images and 100 test images.
We plot the training loss and test accuracy of different methods with the same initialization in Figure~\ref{fig:cifar100}.
From the results, we observe that DSTAdam decreases the training loss faster and smoother than other methods, followed by SGDM and AdaBound, and that Adam is the slowest after the 50th epoch.
For the test part, we find that DSTAdam performs slightly better than AdaBound but outperforms SGDM and Adam.\\

%

\section{Conclusion}
\label{S:7}

In this paper, we have presented a new transition method called DSTAdam,
which achieves a smooth and stable transition from Adam to SGD through scaling operation instead of clipping operation.
Our work extends the research which focus from the bound functions of the AdaBound to the transition operation, provideing a new perspective to understand AdaBound.
Moreover, we have established the $\mathcal{O}({\sqrt T })$ regret bound of the proposed DSTAdam in the online convex setting.
Finally, numerical experiments verify the effectiveness of DSTAdam on the CIFAR-10/100 datasets.
The example results fully show that DSTAdam has fast convergence speed and good generalization performance, which reflect the advantages of Adam and SGD.

\newpage
\bibliographystyle{model1-num-names}
\bibliography{source}


\newpage
\section*{Appendices}
\setcounter{equation}{0}
\renewcommand{\theequation}{A.\arabic{equation}}

\begin{appendices}
\section{Proof of Theorem~\ref{thm:main}}\label{app:A}

We firstly provide the following lemmas for the proof of Theorem~\ref{thm:main}.\\
\begin{lem}\label{lem:A1}\emph{\cite{reddi2019convergence,2002Adaptive}}\label{lem:A1}
Let $a_i\geq 0, i=1,2\cdots, n$. Then
\begin{equation}
\sum\limits_{i = 1}^n {\frac{{{a_i}}}{{\sqrt {\sum\limits_{j = 1}^i {{a_j}} } }}}  \le 2\sqrt {\sum\limits_{i = 1}^n {{a_i}} }.
\end{equation}
\end{lem}
\begin{lem}\label{lem:A2}\emph{\cite{reddi2019convergence,2010Adaptive}}\label{lem:A2}
For any $Q \in \mathcal{S}_ + ^d$ and convex feasible set $\mathcal{F} \subset \mathbb{R}^{d}$,  suppose $u_{1}=  \mathop{\arg\min} _{x \in \mathcal{F}}\left\|Q^{1 / 2}\left(x-z_{1}\right)\right\|$ and $u_{2}=\mathop{\arg\min} _{x \in \mathcal{F}}\left\|Q^{1 / 2}\left(x-z_{2}\right)\right\|$, then $\left\| {{Q^{1/2}}\left( {{u_1} - {u_2}} \right)} \right\| \le \left\| {{Q^{1/2}}\left( {{z_1} - {z_2}} \right)} \right\|$.
\end{lem}

\begin{lem}\label{lem:A3} Suppose that the conditions and parameter setting in Theorem \ref{thm:main} are satisfied, then we have
\[\sum\limits_{t = 1}^T {\frac{1}{{\sqrt t }}{{\left\| {\hat \eta _t^{1/2}{m_t}} \right\|}^2}}  \le \frac{{2\alpha \rho \zeta }}{{{{\left( {1 - {\beta _1}} \right)}^2}}}\sum\limits_{i = 1}^d {{{\left\| {{g_{1:T,i}}} \right\|}_2}}  + \frac{{{r_u}\sqrt {1 + \log T} }}{{{{\left( {1 - {\beta _1}} \right)}^2}}}\sum\limits_{i = 1}^d {{{\left\| {g_{1:T,i}^2} \right\|}_2}} .\]
\end{lem}

\begin{pf} Recall the step~\ref{step:6} in Algorithm~\ref{alg:DSTAdam}, we firstly have
\begin{equation}
\begin{split}
\sum\limits_{t = 1}^T {\frac{1}{{\sqrt t }}{{\left\| {\hat \eta _t^{1/2}{m_t}} \right\|}^2}} & = \sum\limits_{t = 1}^{T - 1} {\frac{1}{{\sqrt t }}{{\left\| {\hat \eta _t^{1/2}{m_t}} \right\|}^2}}  + \frac{1}{{\sqrt T }}\sum\limits_{i = 1}^d {m_{T,i}^2{{\hat \eta }_{T,i}}} \\
& = \sum\limits_{t = 1}^{T - 1} {\frac{1}{{\sqrt t }}{{\left\| {\hat \eta _t^{1/2}{m_t}} \right\|}^2}}  + \frac{1}{{\sqrt T }}\sum\limits_{i = 1}^d {m_{T,i}^2\left( {\left( {\frac{\alpha }{\sqrt{v_{T,i}}} - {r_T}} \right){\rho_T} + {r_T}} \right)} \\
&\le \sum\limits_{t = 1}^{T - 1} {\frac{1}{{\sqrt t }}{{\left\| {\hat \eta _t^{1/2}{m_t}} \right\|}^2}}+ \frac{1}{{\sqrt T }}\sum\limits_{i = 1}^d {m_{T,i}^2\left( \frac{\alpha {\rho_T}}{\sqrt{v_{T,i}}}  + {r_T} \right)}.
\end{split}
\end{equation}
Furthermore, the second term in the right-hand side (RHS) of the above equality can be estimated as follows
\begin{equation}
\begin{split}
&\frac{1}{{\sqrt T }}\sum\limits_{i = 1}^d m_{T,i}^2\left( \frac{\alpha {\rho_T}}{\sqrt{v_{T,i}}}  + {r_T} \right)\\
&\leq\frac{1}{{\sqrt T }}\sum\limits_{i = 1}^d { { {\left( {\sum\limits_{j = 1}^T {\prod\limits_{k = 1}^{T - j} {{\beta _{1( {T - k + 1} )}}} } } \right)\left( {\sum\limits_{j = 1}^T {\prod\limits_{k = 1}^{T - j} {{\beta _{1( {T - k + 1} )}}g_{j,i}^2} } } \right)} \left( \frac{\alpha {\rho_T}}{\sqrt{v_{T,i}}}  + {r_T} \right)} }\\
&\leq\frac{1}{{\sqrt T }}\sum\limits_{i = 1}^d {{\left( {\sum\limits_{j = 1}^T {\beta _1^{T - j}} } \right)\left( {\sum\limits_{j = 1}^T {\beta _1^{T - j}g_{j,i}^2} } \right)\left( \frac{\alpha {\rho _T}}{\sqrt{v_{T,i}}}  + {r_T} \right)} }\\
&\leq\frac{\alpha {\rho_T}}{(1-\beta_1){\sqrt T }}\sum\limits_{i = 1}^d {{\left( {\sum\limits_{j = 1}^T {\beta _1^{T - j}g_{j,i}^2} } \right) \frac{1}{\sqrt{v_{T,i}}}} }+\frac{{r_T}}{(1-\beta_1){\sqrt T }}\sum\limits_{i = 1}^d {{\left( {\sum\limits_{j = 1}^T {\beta _1^{T - j}g_{j,i}^2} } \right)  } }.
\end{split}
\end{equation}
By condition \textbf{(C1)} in Theorem~\ref{thm:main}, the first term in the RHS of the above inequality can be further simplified as
\begin{equation}
\begin{split}
\frac{\alpha {\rho_T}}{(1-\beta_1){\sqrt T }}\sum\limits_{i = 1}^d {{\left( {\sum\limits_{j = 1}^T {\beta _1^{T - j}g_{j,i}^2} } \right) \frac{1}{\sqrt{v_{T,i}}}} }  &\leq\frac{\alpha {\rho_T}}{(1-\beta_1)}\sum\limits_{i = 1}^d {{\left( {\sum\limits_{j = 1}^T {\beta _1^{T - j}g_{j,i}^2} } \right)} }{\frac{1}{\frac{1} {\zeta} {\sqrt {\sum\limits_{j = 1}^T {g_{j,i}^2} } }}} \\
&\leq\frac{{\alpha {\rho_T}\zeta }}{{\left( {1 - {\beta _1}} \right)}}\sum\limits_{i = 1}^d { {\sum\limits_{j = 1}^T {\frac{{\beta _1^{T - j}g_{j,i}^2}}{{\sqrt {\sum\limits_{k = 1}^j {g_{k,i}^2} } }}} }}.
\end{split}
\end{equation}
Combining the above estimates, we arrive at
\begin{equation}
\begin{split}
\sum\limits_{t = 1}^T {\frac{1}{{\sqrt t }}{{\left\| {\hat \eta _t^{1/2}{m_t}} \right\|}^2}}&\le \sum\limits_{t = 1}^{T - 1} {\frac{1}{{\sqrt t }}{{\left\| {\hat \eta _t^{1/2}{m_t}} \right\|}^2}}  \\
&\quad+ \frac{{\alpha {\rho_T}\zeta }}{{\left( {1 - {\beta _1}} \right)}}\sum\limits_{i = 1}^d { {\sum\limits_{j = 1}^T {\frac{{\beta _1^{T - j}g_{j,i}^2}}{{\sqrt {\sum\limits_{k = 1}^j {g_{k,i}^2} } }}} } }  + \frac{{{r_T}}}{{\left( {1 - {\beta _1}} \right)\sqrt T }}\sum\limits_{i = 1}^d {\sum\limits_{j = 1}^T {\beta _1^{T - j}g_{j,i}^2} } .
\end{split}
\end{equation}
By the principle of mathematical induction and the double-sum trick, we have
\begin{equation}
\begin{split}
\sum\limits_{t = 1}^T {\frac{1}{{\sqrt t }}{{\left\| {\hat \eta _t^{1/2}{m_t}} \right\|}^2}} & \le \sum\limits_{t = 1}^T {\frac{{\alpha {\rho_t}\zeta }}{{\left( {1 - {\beta _1}} \right)}}\sum\limits_{i = 1}^d { {\sum\limits_{j = 1}^t {\frac{{\beta _1^{t - j}g_{j,i}^2}}{{\sqrt {\sum\limits_{k = 1}^j {g_{k,i}^2} } }}} } } }  + \sum\limits_{t = 1}^T {\frac{{{r_t}}}{{\left( {1 - {\beta _1}} \right)\sqrt t }}\sum\limits_{i = 1}^d {\sum\limits_{j = 1}^t {\beta _1^{t - j}g_{j,i}^2} } } \\
 &\le \frac{{\alpha\rho \zeta }}{{\left( {1 - {\beta _1}} \right)}}\sum\limits_{i = 1}^d {\sum\limits_{t = 1}^T {\sum\limits_{j = 1}^t {\frac{{\beta _1^{t - j}g_{j,i}^2}}{{\sqrt {\sum\limits_{k = 1}^j {g_{k,i}^2} } }}} } }  + \frac{{{r_u}}}{{\left( {1 - {\beta _1}} \right)}}\sum\limits_{i = 1}^d {\sum\limits_{t = 1}^T {\frac{{\rm{1}}}{{\sqrt t }}} \sum\limits_{j = 1}^t {\beta _1^{t - j}} g_{j,i}^2} \\
& = \frac{{\alpha \rho \zeta }}{{\left( {1 - {\beta _1}} \right)}}\sum\limits_{i = 1}^d {\sum\limits_{t = 1}^T {\frac{{g_{t,i}^2}}{{\sqrt {\sum\limits_{k = 1}^t {g_{k,i}^2} } }}\sum\limits_{j = t}^T {\beta _1^{j - t}} } }  + \frac{{{r_u}}}{{\left( {1 - {\beta _1}} \right)}}\sum\limits_{i = 1}^d {\sum\limits_{t = 1}^T {g_{t,i}^2} \sum\limits_{j = t}^T {\frac{{\beta _1^{j - t}}}{{\sqrt j }}} }\\
& \le \frac{{\alpha \rho\zeta }}{{{{\left( {1 - {\beta _1}} \right)}^2}}}\sum\limits_{i = 1}^d {\sum\limits_{t = 1}^T {\frac{{g_{t,i}^2}}{{\sqrt {\sum\limits_{k = 1}^t {g_{k,i}^2} } }}} }  + \frac{{{r_u}}}{{\left( {1 - {\beta _1}} \right)^2}}\sum\limits_{i = 1}^d {\sum\limits_{t = 1}^T \frac{g_{t,i}^2}{{\sqrt t }} }.
\end{split}
\end{equation}
Upon applying Lemma~\ref{lem:A1} and using Cauchy-Schwarz inequality, we obtain
\begin{equation}
\begin{split}
\sum\limits_{t = 1}^T {\frac{1}{{\sqrt t }}{{\left\| {\hat \eta _t^{1/2}{m_t}} \right\|}^2}}
 &\le \frac{{\alpha\rho \zeta }}{{{{\left( {1 - {\beta _1}} \right)}^2}}}\sum\limits_{i = 1}^d {2\sqrt {\sum\limits_{t = 1}^T {g_{t,i}^2} } }  + \frac{{{r_u}}}{{{{\left( {1 - {\beta _1}} \right)}^2}}}\sum\limits_{i = 1}^d {{{\left\| {g_{1:T,i}^2} \right\|}_2}\sqrt {\sum\limits_{t = 1}^T {\frac{1}{t}} } }  \\
& \le \frac{{2\alpha \rho \zeta }}{{{{\left( {1 - {\beta _1}} \right)}^2}}}\sum\limits_{i = 1}^d {{{\left\| {{g_{1:T,i}}} \right\|}_2}}  + \frac{{{r_u}\sqrt {1 + \log T} }}{{{{\left( {1 - {\beta _1}} \right)}^2}}}\sum\limits_{i = 1}^d {{{\left\| {g_{1:T,i}^2} \right\|}_2}}.
\end{split}
\end{equation}
\end{pf}

Now we are ready to prove Theorem~\ref{thm:main}.

\noindent\textbf{Proof of Theorem 1}. By the definition of projection, we have
\begin{equation}
\begin{split}
{\theta _{t + 1}} = {\Pi _{{\rm{{\cal F}}}, {\hat \eta _t^{ - 1}} }}\left( {{\theta _t} - \frac{1}{{\sqrt t }}{{\hat \eta }_t}  {m_t}} \right) = \mathop {\mathop{\arg\min} }\limits_{\theta  \in {\rm{{\cal F}}}} \left\| {\hat \eta _t^{ - 1/2}  \left( {\theta  - \left( {{\theta _t} - \frac{1}{{\sqrt t }}{{\hat \eta }_t} {m_t}} \right)} \right)} \right\|.
\end{split}
\end{equation}
Note that $\Pi_{\mathcal{F}, \hat{\eta}_{t}^{-1}}\left({\theta ^*}\right) = {\theta ^*}$ for all $\theta^{*} \in \mathcal{F}$. Using Lemma~\ref{lem:A2} with ${u_1} = {\theta _{t + 1}}$ and ${u_2} = {\theta ^*}$, we have
\begin{equation}
\begin{split}
{\left\| {\hat \eta _t^{ - 1/2}\left( {{\theta _{t + 1}} - {\theta ^*}} \right)} \right\|^2} &\le {\left\| {\hat \eta _t^{ - 1/2}\left( {{\theta _t} - \frac{1}{{\sqrt t }}{{\hat \eta }_t}{m_t} - {\theta ^*}} \right)} \right\|^2}\\
& = {\left\| {\hat \eta _t^{ - 1/2}\left( {{\theta _t} - {\theta ^*}} \right)} \right\|^2} + \frac{1}{t}{\left\| {\hat \eta _t^{1/2}{m_t}} \right\|^2} - \frac{2}{{\sqrt t }}\left\langle {{\beta _{1t}}{m_{t - 1}} + \left( {1 - {\beta _{1t}}} \right){g_t},{\theta _t} - {\theta ^*}} \right\rangle.
\end{split}
\end{equation}
Rearranging the terms in the above inequality, we have
\begin{equation}
\begin{split}
 &\left\langle g_{t}, \theta_{t}-\theta^{*}\right\rangle \\
 &\leq \frac{\sqrt{t}}{2\left(1-\beta_{1 t}\right)}\left[\left\|\hat{\eta}_{t}^{-1 / 2}\left(\theta_{t}-\theta^{*}\right)\right\|^{2}-\left\|\hat{\eta}_{t}^{-1 / 2}\left(\theta_{t+1}-\theta^{*}\right)\right\|^{2}\right] \\
&\quad+\frac{1}{2 \sqrt{t}\left(1-\beta_{1 t}\right)}\left\|\hat{\eta}_{t}^{1 / 2} m_{t}\right\|^{2}-\frac{\beta_{1 t}}{\left(1-\beta_{1 t}\right)}\left\langle m_{t-1}, \theta_{t}-\theta^{*}\right\rangle \\
&\leq  \frac{\sqrt{t}}{2\left(1-\beta_{1 t}\right)}\left[\left\|\hat{\eta}_{t}^{-1 / 2}\left(\theta_{t}-\theta^{*}\right)\right\|^{2}-\left\|\hat{\eta}_{t}^{-1 / 2}\left(\theta_{t+1}-\theta^{*}\right)\right\|^{2}\right]+\frac{1}{2 \sqrt{t}\left(1-\beta_{1 t}\right)}\left\|\hat{\eta}_{t}^{1 / 2} m_{t}\right\|^{2} \\
&\quad+\frac{\beta_{1 t} \sqrt{t}}{2\left(1-\beta_{1 t}\right)}\left\|\hat{\eta}_{t}^{-1 / 2}\left(\theta_{t}-\theta^{*}\right)\right\|^{2}+\frac{\beta_{1 t}}{2 \sqrt{t}\left(1-\beta_{1 t}\right)}\left\|\hat{\eta}_{t}^{1 / 2} m_{t-1}\right\|^{2},
\end{split}
\end{equation}
where the last inequality follows from a basic inequality $xy\leq\frac{\kappa x^2}{2}+\frac{y^2}{2\kappa}, \kappa>0$. By the convexity of function ${f_t}$ and the above inequality, we arrive at
\begin{equation}
\label{eq:ffconv}
\begin{split}
&\sum\limits_{t = 1}^T {\left( {{f_t}\left( {{\theta _t}} \right) - {f_t}\left( {{\theta ^*}} \right)} \right)} \leq \sum\limits_{t = 1}^T \left\langle g_{t}, \theta_{t}-\theta^{*}\right\rangle\\
& \le \sum\limits_{t = 1}^T {\frac{{\sqrt t }}{{2\left( {1 - {\beta _{1}}} \right)}}\left[ {{{\| {\hat \eta _t^{ - 1/2}\left( {{\theta _t} - {\theta ^*}} \right)} \|}^2} \!-\! {{\| {\hat \eta _t^{ - 1/2}\left( {{\theta _{t + 1}} - {\theta ^*}} \right)} \|}^2}} \right]} \! + \!\sum\limits_{t = 1}^T {\frac{{{\beta _{1t}}\sqrt t }}{{2\left( {1 - {\beta _{1t}}} \right)}}{{\left\| {\hat \eta _t^{ - 1/2}\left( {{\theta _t} - {\theta ^*}} \right)} \right\|}^2}} \\
&\quad+ \sum\limits_{t = 1}^T {\frac{1}{{2\sqrt t \left( {1 - {\beta _1}} \right)}}{{\left\| {\hat \eta _t^{1/2}{m_t}} \right\|}^2}}  + \sum\limits_{t = 1}^{T - 1} {\frac{{{\beta _1}}}{{2\sqrt {t + 1} \left( {1 - {\beta _1}} \right)}}{{\left\| {\hat \eta _{t + 1}^{1/2}{m_t}} \right\|}^2}}.
\end{split}
\end{equation}
By the conditions \textbf{(C2)} and $\left\|x-y\right\|_{\infty} \leq D_{\infty}$, $\forall x,y\in \mathcal F$. Then, the first term in the RHS of \eqref{eq:ffconv} can be estimated as
\begin{equation}
\begin{split}
&\sum_{t=1}^{T} \frac{\sqrt{t}}{2\left(1-\beta_{1 t}\right)}\left[\left\|\hat{\eta}_{t}^{-1 / 2}\left(\theta_{t}-\theta^{*}\right)\right\|^{2}-\left\|\hat{\eta}_{t}^{-1 / 2}\left(\theta_{t+1}-\theta^{*}\right)\right\|^{2}\right] \\
&\leq \frac{\left\|\hat{\eta}_{1}^{-1 / 2}\left(\theta_{1}-\theta^{*}\right)\right\|}{2\left(1-\beta_{1}\right)}+\sum_{t=2}^{T} \frac{\sqrt{t}}{2\left(1-\beta_{1 t}\right)}\left[\left\|\hat{\eta}^{-1 / 2}\left(\theta_{t}-\theta^{*}\right)\right\|^{2}-\left\|\hat{\eta}_{t-1}^{-1 / 2}\left(\theta_{t}-\theta^{*}\right)\right\|^{2}\right] \\
&\leq \frac{\left\|\hat{\eta}_{1}^{-1 / 2} \right\|}{2\left(1-\beta_{1}\right)}D_{\infty}^2+\frac{1}{2\left(1-\beta_{1}\right)} \sum_{t=2}^{T}\left(\sqrt{t}\left\|\hat{\eta}_{t}^{-1 / 2} D_{\infty}\right\|^{2}-\sqrt{t-1}\left\|\hat{\eta}_{t-1}^{-1 / 2} D_{\infty}\right\|^{2}\right) \\
&=\frac{\sqrt{T}D_{\infty}^2\left\|\hat{\eta}_{T}^{-1 / 2} \right\|^{2}}{2\left(1-\beta_{1}\right)}.
\end{split}
\end{equation}
By condition \textbf{(C2)} in Theorem~\ref{thm:main}, the last term in the RHS of \eqref{eq:ffconv} can be simplified as
\begin{equation}
\begin{split}
\sum\limits_{t = 1}^{T - 1} {\frac{{{\beta _1}}}{{2\sqrt {t + 1} \left( {1 - {\beta _1}} \right)}}{{\left\| {\hat \eta _{t + 1}^{1/2}{m_t}} \right\|}^2}}
 \leq \sum\limits_{t = 1}^{T} {\frac{1}{{2\sqrt t \left( {1 - {\beta _1}} \right)}}{{\left\| {\hat \eta _t^{1/2}{m_t}} \right\|}^2}}.
\end{split}
\end{equation}
Summarizing the above estimates and applying Lemma \ref{lem:A3}, we obtain
\begin{equation}
\begin{split}
\sum\limits_{t = 1}^T {\left( {{f_t}\left( {{\theta _t}} \right) - {f_t}\left( {{\theta ^*}} \right)} \right)}
&\leq \frac{\sqrt{T}D_{\infty}^2\left\|\hat{\eta}_{T}^{-1 / 2} \right\|^{2}}{2\left(1-\beta_{1}\right)}  + \sum\limits_{t = 1}^T {\frac{{{\beta _{1t}}\sqrt t }}{{2\left( {1 - {\beta _{1t}}} \right)}}{{\left\| {\hat \eta _t^{ - 1/2}\left( {{\theta _t} - {\theta ^*}} \right)} \right\|}^2}} \\
&\quad +\frac{{2\alpha \rho \zeta }}{{{{\left( {1 - {\beta _1}} \right)}^3}}}\sum\limits_{i = 1}^d {{{\left\| {{g_{1:T,i}}} \right\|}_2}}  + \frac{{{r_u}\sqrt {1 + \log T} }}{{{{\left( {1 - {\beta _1}} \right)}^3}}}\sum\limits_{i = 1}^d {{{\left\| {g_{1:T,i}^2} \right\|}_2}}\\
&\leq \frac{\sqrt{T} D_{\infty}^{2}}{2\left(1-\beta_{1}\right)} \sum_{i=1}^{d} \hat{\eta}_{T, i}^{-1}+\frac{D_{\infty}^{2}}{2\left(1-\beta_{1}\right)} \sum_{t=1}^{T} \sum_{i=1}^{d} \frac{\beta_{1 t} \sqrt{t}}{\hat{\eta}_{t, i}} \\
&\quad +\frac{2 \alpha \rho \zeta}{\left(1-\beta_{1}\right)^{3}} \sum_{i=1}^{d}\left\|g_{1: T, i}\right\|_{2}+\frac{r_{u} \sqrt{1+\log T}}{\left(1-\beta_{1}\right)^{3}} \sum_{i=1}^{d}\left\|g_{1: T, i}^{2}\right\|_{2} .
\end{split}
\end{equation}

We proceed to prove Corollary~\ref{cor:4.1} and Corollary~\ref{cor:4.2} from Theorem~\ref{thm:main}.

\noindent\textbf{Proof of Corollary 1}. Recall the step~\ref{step:6} in Algorithm~\ref{alg:DSTAdam}, we have
\begin{equation}\label{eq:etbound}
\begin{split}
\hat{\eta}_{t, i}^{-1} &=\left(\rho_{t}\left(\frac{\alpha}{\sqrt{v_{t, i}}}-r_{t}\right)+r_{t}\right)^{-1}\\
&=\frac{1}{\frac{\alpha \rho_{t}}{\sqrt{v_{t,i}}}-r_{t} \rho_{t}+r_{t}} \\
& \leq \frac{1}{r_{t}-r_{t} \rho_{t}} \\
& \leq \frac{1}{r_{l}(1-\rho)}.
\end{split}
\end{equation}
Since ${\beta _{1t}} = {\beta _1}{\lambda ^{t - 1}}$, it follows that
\begin{equation}
\begin{split}
\sum_{t=1}^{T} \sum_{i=1}^{d} \frac{\beta_{1 t} \sqrt{t}}{\hat{\eta}_{t, i}} & \leq \frac{\beta_{1} d}{r_{l}(1-\rho)} \sum_{t=1}^{T} \lambda^{t-1} \sqrt{t} \\
& \leq \frac{d}{r_{l}(1-\rho)} \sum_{t=1}^{T} \lambda^{t-1} t \\
&=\frac{d}{r_{l}(1-\rho)}\left(\frac{1-\lambda^{T}}{(1-\lambda)^{2}}-\frac{\lambda^{T} T}{(1-\lambda)}\right) \\
& \leq \frac{d}{r_{l}(1-\rho)(1-\lambda)^{2}}.
\end{split}
\end{equation}
Based on the result in Theorem~\ref{thm:main}, we obtain
\begin{equation}
\begin{split}
R(T) & \leq \frac{\sqrt{T} D_{\infty}^{2}}{2\left(1-\beta_{1}\right)} \sum_{i=1}^{d} \hat{\eta}_{T, i}^{-1}+\frac{d D_{\infty}^{2}}{2 r_{l}(1-\rho)(1-\lambda)^{2}\left(1-\beta_{1}\right)} \\
&\quad+\frac{2 \alpha \rho \zeta}{\left(1-\beta_{1}\right)^{3}} \sum_{i=1}^{d}\left\|g_{1: T, i}\right\|_{2}+\frac{r_{u} \sqrt{1+\log T}}{\left(1-\beta_{1}\right)^{3}} \sum_{i=1}^{d}\left\|g_{1: T, i}^{2}\right\|_{2}.
\end{split}
\end{equation}

\noindent\textbf{Proof of Corollary 2}.
Since  ${\beta _{1t}} = {\beta _1}/t$, it follows that
\begin{equation}
\begin{split}
\sum_{t=1}^{T} \sum_{i=1}^{d} \frac{\beta_{1 t} \sqrt{t}}{\hat{\eta}_{t, i}} & \leq \frac{\beta_{1} d}{r_{l}(1-\rho)} \sum_{t=1}^{T} \frac{\sqrt{t}}{t} \\
& \leq \frac{d}{r_{l}(1-\rho)} \sum_{t=1}^{T} 2(\sqrt{t}-\sqrt{t-1}) \\
& \leq \frac{2 d \sqrt{T}}{r_{l}(1-\rho)}.
\end{split}
\end{equation}
Based on the result in Theorem~\ref{thm:main}, we obtain
\begin{equation}
\begin{split}
R(T) &\leq \frac{\sqrt{T} D_{\infty}^{2}}{2\left(1-\beta_{1}\right)} \sum_{i=1}^{d} \hat{\eta}_{T, i}^{-1}+\frac{d D_{\infty}^{2} \sqrt{T}}{r_{l}(1-\rho)\left(1-\beta_{1}\right)} \\
&\quad +\frac{2 \alpha \rho \zeta}{\left(1-\beta_{1}\right)^{3}} \sum_{i=1}^{d}\left\|g_{1: T, i}\right\|_{2}+\frac{r_{u} \sqrt{1+\log T}}{\left(1-\beta_{1}\right)^{3}} \sum_{i=1}^{d}\left\|g_{1: T, i}^{2}\right\|_{2} .
\end{split}
\end{equation}
\end{appendices}










\end{document}